\documentclass[lettersize,journal]{IEEEtran}
\usepackage{amsmath,amsfonts}
\usepackage{algorithm}
\usepackage{array}
\usepackage[caption=false,font=normalsize,labelfont=sf,textfont=sf]{subfig}
\usepackage{textcomp}
\usepackage{stfloats}
\usepackage{url}
\usepackage{verbatim}
\usepackage{graphicx}
\usepackage{cite}
\usepackage{framed,multirow}
\usepackage{lineno,hyperref}

\usepackage{amssymb}

\usepackage{latexsym}

\usepackage{algpseudocode}
\usepackage{booktabs}
\usepackage[normalem]{ulem}
\usepackage{dutchcal}
\begin{document}

\title{Few-shot Image Generation via Information Transfer from the Built Geodesic Surface}


\author{Yuexing~Han,~Liheng~Ruan,~and~Bing~Wang%
\thanks{Y. Han is with the School of Computer Engineering and Science, Shanghai University, 99 Shangda Road, Shanghai 200444, China, Zhejiang Laboratory, Hangzhou 311100, China and Key Laboratory of Silicate Cultural Relics Conservation (Shanghai University), Ministry of Education (e-mail: Han\_yx@i.shu.edu.cn).}
\thanks{L. Ruan and B. Wang are with the School of Computer Engineering and Science, Shanghai University, 99 Shangda Road, Shanghai 200444, China.}}

\maketitle

\begin{abstract}
  Images generated by most of generative models trained with limited data often exhibit deficiencies in either fidelity, diversity, or both. One effective solution to address the limitation is few-shot generative model adaption. However, the type of approaches typically rely on a large-scale pre-trained model, serving as a source domain, to facilitate information transfer to the target domain. In this paper, we propose a method called Information Transfer from the Built Geodesic Surface (ITBGS), which contains two modules: Feature Augmentation on Geodesic Surface (FAGS); Interpolation and Regularization (I\&R). With the FAGS module, a pseudo-source domain is created by projecting image features from the training dataset into the Pre-Shape Space, subsequently generating new features on the Geodesic surface. Thus, no pre-trained models is needed for the adaption process during the training of generative models with FAGS. I\&R module are introduced for supervising the interpolated images and regularizing their relative distances, respectively, to further enhance the quality of generated images. Through qualitative and quantitative experiments, we demonstrate that the proposed method consistently achieves optimal or comparable results across a diverse range of semantically distinct datasets, even in extremely few-shot scenarios.
\end{abstract}


\section{Introduction}
\label{sec:Introduction}

Most of the image generation methods such as Variational Auto-encoders (VAEs)\cite{kingma2013auto}, Generative Adversarial Networks (GANs) \cite{goodfellow_generative_2020}, and Diffusion models \cite{ho2020denoising}, have demonstrated their ability to produce images with a satisfactory combination of fidelity and diversity. Despite their impressive performance, the demanding of large-scale image datasets pose a considerable challenge for training the image generation methods. However, some certain domains, such as medical, remote sensing, and material images, pose challenges in terms of data acquisition, making it difficult to obtain the large-scale datasets typically required for training \cite{shorten2019survey,han2023data}. The scarcity of data hinders common downstream tasks like target detection, image classification, semantic segmentation, and so on. Generating images with image generation models serves as a way to effectively expand the dataset for downstream tasks. In such scenarios, the goal of image generation is to maximize the utility of a limited dataset by training a model capable of producing images that are both high in fidelity and diversity \cite{han2023data}. 

In recent years, there have also been some studies for image generation under few-shot setting. These generative methods can be broadly categorized into two types. The first type of methods is the few-shot generative model adaption \cite{ojha_few-shot_2021,xiao_few_2022}, which leverages semantically relevant pre-trained image generation models in the task setup. The type of methods involves acquiring additional data to form a source domain and transferring the rich image information from the source domain to the target generator during training. It is worth noting that the scope of the adaption approach is constrained by the need for a strong semantic correlation between the source and target domain \cite{ojha_few-shot_2021}. That is to say, semantically irrelevant image information cannot be effectively transferred. Consequently, while the type of methods can be applied even in extremely few-shot scenarios, e.g, less equal than 10 samples, the source generator often necessitates numerous samples to pre-train.

The second type of methods is training models from scratch without utilizing source domain \cite{karras2020training,liu_towards_2021,kong_few-shot_2022}. Although these methods tend to exhibit improved performance under low-shot conditions, their effectiveness diminishes when confronted with more extreme few-shot scenarios. In such cases, the type of these methods are more susceptible to issues like overfitting and ``stairlike" phenomenon \cite{kong_few-shot_2022}. Some methods, such as MixDL \cite{kong_few-shot_2022} have demonstrated the capacity to produce favorable output even in the challenging extreme few-shot scenarios, e.g., 10 samples. However, their improvement come at the expense of reduced fidelity.

Finding the delicate balance between fidelity and diversity remains the top challenge in the field of extreme few-shot image generation. The type of the generative model adaption methods have exhibited noteworthy achievements in few-shot image generation methods \cite{ojha_few-shot_2021,xiao_few_2022,li_few-shot_2020}. However, acquiring an appropriate pre-trained model to serve as the source domain can be difficult in many cases. The absence of a pre-trained generator implies the absence of a readily available source domain for information transfer. 

\begin{figure*}[htbp]
  \begin{center}
  \includegraphics[width=0.9\linewidth]{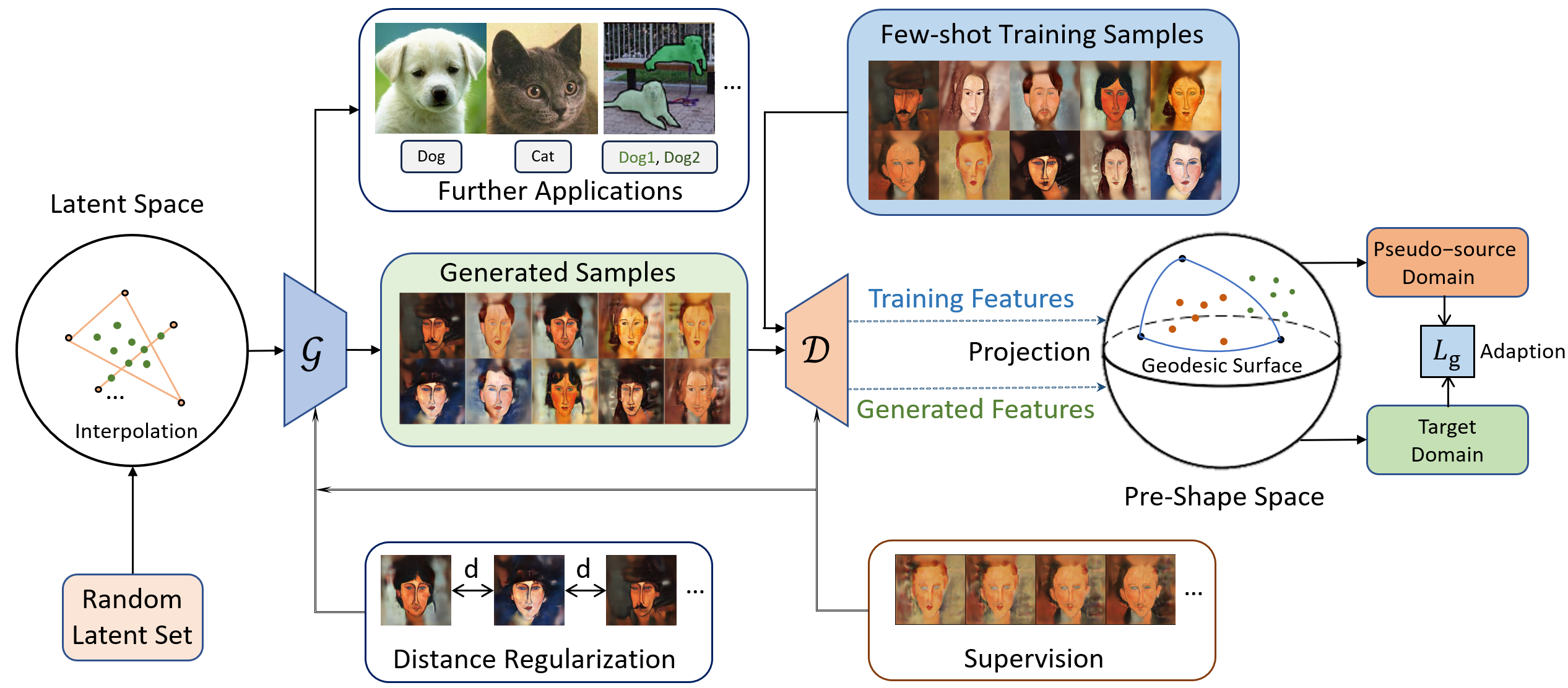}
  \end{center}\caption{Our motivation on Information Transfer from the Built Geodesic Surface (ITBGS). A pseudo-source domain is created by manifold data augmenting the features extracted only from extremely few training samples, e.g., 10 samples, and adapt to the target domain for training generator in the Pre-Shape Space. We interpolate the latents within the target domain, ensuring that the generated features maintains a similar spatial distribution to the augmented features. The adaption method is achieved by aligning the inherent structural information of the two aforementioned features. Additionally, the interpolation and regularization strategies are employed to the generated samples and features. The trained generator can be used for further applications, such as few-shot image classification and instance segmentation.} 
  \label{fig:motivation}
  \end{figure*}

To overcome the aforementioned shortcomings, we design a method called Information Transfer from the Built Geodesic Surface (ITBGS), which consists of Feature Augmentation on geodesic Surface (FAGS) module and Interpolation and Regularization (I\&R) module. Figure \ref{fig:motivation} shows our motivation. FAGS focuses on the creation of a pseudo-source domain using the available few-shot training samples. Recognizing the effectiveness of data augmentation techniques in few-shot tasks \cite{qin2020diversity,zhou2021flipda,osahor2022ortho}, we adopt the idea of data augmentation to generate the pseudo-source domain. Instead of relying on conventional data augmentation methods like horizontal or vertical flipping, brightness adjustments and color modifications, we turn to data augmentation method rooted in the Shape Space theory \cite{han2010recognition,han2014recognizing}. Instead of constructing a optimal Geodesic curve \cite{han2023fagcfeature}, new data are obtained by building a Geodesic surface in the Pre-Shape Space with the training samples. Subsequently, we aim to transfer the inherent image information embedded in the pseudo-source domain to the target generator.

With the help of FAGS, the generated samples maintain better quality. However, blurriness occurs in the intermediate interpolated samples. We further propose the I\&R module to supervise and regularize the relative distances of the interpolated samples. The model trained with ITBGS can be used for augmenting the training set in further applications, such as few-shot image classification and instance segmentation.

We show both qualitative and quantitative experimental results in our paper, illustrating the balance our model attains in terms of fidelity and diversity within the generated images. The proposed ITBGS produces commendable results across diverse 10-shot datasets. Its ability in generating realistic facial images is also demonstrated. 

In summary, our key contributions can be outlined as follows:
\begin{itemize}
  \item A pseudo-source domain is created for transferring the information to the target generator where no large-scale dataset or pre-trained generator exists.
  \item The Shape Space theory is introduced to build a Geodesic surface in the Pre-Shape Space for augmenting data.
  \item Interpolation and regularization strategies are employed to enhance both image quality and diversity.
\end{itemize}

\section{Related Work}
\label{sec:Related Work}

\subsection{Few-shot Image Generation}
\label{sec:Few-shot Image Generation}
As in the conventional GAN training procedure, a dataset $\mathbb{D}_{real}$ is given to train the generator $\mathcal{G}$. The noise or latent vectors $z\sim p(z) \subset \mathbb{R}^d$ is drawn from the $\mathcal{D}$-dimensional latent space for better properties if using StyleGAN2  \cite{karras2019style,zhu2020improved,alaluf2021restyle}, rather than from the Gaussian distribution. $\mathcal{G}$ maps latent vectors $z$ into generated images $\mathcal{G}(z)$ in the pixel space. The optimization function for generator $\mathcal{G}$ and a learned discriminator $\mathcal{D}$ are $L^\mathcal{G}_{adv}$ and $L^\mathcal{D}_{adv}$, respectively, defined as follows \cite{goodfellow_generative_2020}:

\begin{equation} \label{eq:Lg_adv}
  L^\mathcal{G}_{adv}=-\mathbb{E}_{z\sim p(z)}[log(\mathcal{D}(\mathcal{G}(z)))],
\end{equation}
\noindent and
\begin{equation} \label{eq:Ld_adv}
  L^\mathcal{D}_{adv}=\mathbb{E}_{x\sim \mathbb{D}_{real}}[log(1-\mathcal{D}(x))]+\mathbb{E}_{z\sim p(z)}[log(\mathcal{D}(\mathcal{G}(z)))].
\end{equation}

In few-shot scenarios, common image generation methods are highly susceptible to cause overfitting or memorization. The primary solutions to address the problem of few-shot scenarios can be categorized into two categories of distinct approaches:

One category of approaches is few-shot image generation with source domain, which transfer the rich image information from the source domain to the target generator, similar to transfer learning \cite{ojha_few-shot_2021,xiao_few_2022,li_few-shot_2020}. Source domain is obtained by the pre-trained image generation models, which are originated from a much larger auxiliary dataset with semantic relevance to the training set. Ojha et al. \cite{ojha_few-shot_2021} introduced a cross-domain distance consistency loss, ensuring the similarity distribution of the generated images by the target generator closely aligns with the source domain during training. The alignment aims to enhance the diversity of the generated images. Building upon the foundation, RSSA \cite{xiao_few_2022} further transferred the inherent image structure information through self-correlation consistency loss and disturbance correlation consistency loss. These loss functions effectively address potential issues such as identity degradation and image distortion.

The other is few-shot image generation without source domain, which focuses on avoiding the need for additional information. Data augmentation serves as a notable solution in the context. The techniques yield impressive generation outcomes by enhancing diversity and reducing the risk of overfitting. The approaches such as DiffAugment \cite{zhao_differentiable_2020} and ADA \cite{karras2020training} effectively expanded the number of real and fake samples to prevent the overfitting issue. An alternative approach involves directly altering the architecture of the generative model. For instance, a skip-layer excitation module to the generator and the discriminator is updated by a self-supervised training scheme in FastGAN \cite{liu_towards_2021}. The introduced module not only accelerates training speed and improves stability, but also facilitates the rapid and efficient generation of high-resolution images. Some methods like SinGAN \cite{shaham_singan_2019} and CoSinGAN \cite{hinz_improved_2021} include an extreme scenario, focusing on one-shot image generation. Han et al. \cite{gur2020hierarchical,han2023data} proposed an improved HP-VAE-GAN to generate material images for data augmentation. However, the improved HP-VAE-GAN lacks effectiveness for the non textured images. Most of the category of approaches struggle to achieve satisfactory output in extremely few-shot scenarios, e.g., less equal than 10 samples. Among the extreme few-shot image generation approaches, MixDL \cite{kong_few-shot_2022} stands as the single approach that utilizes no additional data or pre-trained models during the training. MixDL notably enhances the diversity of generated images, albeit with a comparatively slight compromise in fidelity.



\subsection{Feature Augmentation}
\label{sec:Feature Augmentation}

When traversing along the feature space, it is more likely to encounter realistic samples compared to the input space \cite{devries2017dataset}. Feature augmentation manipulates feature vectors, rather than augments only on the image level \cite{zhang_mixup_2018}. Some methods performed simple operations on features extracted by neural networks, such as adding noise \cite{devries2017dataset} and linear combination \cite{verma2019manifold}. More complex transformations are also proposed for feature augmentation. For instance, a learned refinement and augmentation method is introduced in FeatMatch, which use information from prototypical class representations \cite{kuo2020featmatch}. MixStyle mixed the feature statistics of two
instances with a random convex weight to generate new styles \cite{zhou2021domain}. Instead of directly obtaining features, Mangla et al. \cite{mangla2020charting} leveraged self-supervision to obtain a suitable feature manifold before applying manifold mixup in their training procedure. Similarly, Khan et al. \cite{khan2020post} generated new samples by learning a generative model over both low-level and high-level deep feature spaces. Han et al. \cite{han2023fagcfeature} constructed Geodesic curve using features extracted by a pre-trained ViT, and obtained new features from the built Geodesic curves. In our method, we extract features from the continuous updating discriminator and build Geodesic surface during the every epoch of training procedure for feature augmentation.

\subsection{The Shape Space Theory}
\label{sec:Shape Space}

The Shape Space theory, originally introduced by Kendall in 1984 \cite{kendall_shape_1984}, has been a foundational concept in geometric data analysis. The Shape Space theory defines shape as the geometric information that persists when positional, scaling, and rotational effects are removed. 

In recent years, some interesting combination among the Shape Space theory and other domains were proposed. For instance, Kilian et al. \cite{kilian2007geometric} presented continuous deformation of 3D models using Geodesic interpolation in the Shape Space. Han et al. \cite{han2010recognition,han2014recognizing} proposed a object recognition method using the Shape Space theory. They projected object contours and identified a Geodesic curve that aligns with the diverse potential shapes of the given object type in the Pre-Shape Space. Similarly, Paskin et al. \cite{paskin2022kendall} projected 3D shark bone landmarks into the Shape Space and inferred the 3D pose of the shark within the 2D image on a Geodesic surface. Notably, Friji et al. \cite{friji2021geometric} combined the Shape Space theory with deep learning, achieving state-of-the-art outcomes in human pose recognition tasks. However, there is currently no other work that combines the Shape Space theory with few-shot image generation.

In a two-dimensional Euclidean space, a shape $P$ can be represented through a set of landmarks, specifically defined as $P=\{p_1(x_1, y_1), ..., p_m(x_m, y_m)\} \in \mathbb{R}^{2\times m}$. However, the process of projecting $P$ into the Shape Space involves complicated operations within the complex domain. Thus, the majority of research focuses on the Pre-Shape Space. The projection of $P$ into the Pre-Shape Space is achieved by a mean-reduction operation $\mathcal{Q}(\cdot)$ and normalization operation $\mathcal{V}(\cdot)$, leading to the Pre-Shape $\tau$:

\begin{equation}
  P'=\mathcal{Q}(P)=\{p'_i=(x_i-\bar{x}, y_i-\bar{y})\},
  \label{eq:mean-reduction}
\end{equation}
\noindent and
\begin{equation}
  \tau=\mathcal{V}(\mathcal{Q}(P))=\mathcal{V}(P') = \frac{P'}{\|P'\|},
  \label{eq:normalization}
\end{equation}

\noindent where $i=1,...,m$ and $m$ denotes the number of landmarks. $\|\cdot\|$ denotes the Euclidean norm.

The Pre-Shape Space can be conceptualized as a hypersphere, where a point on the hypersphere is achieved through the projection outlined in Formula \ref{eq:mean-reduction} and \ref{eq:normalization}. Pre-Shapes retain rotational effects in contrast to shapes.

Han et al. proposed some methods to generate more new Pre-Shapes from two or three samples on the Geodesic curve or surface \cite{han2010recognition,han2014recognizing}. The Geodesic curve is derived from the following formula when provided two Pre-Shapes $\tau_1$ and $\tau_2$ in the Pre-Shape Space:

\begin{equation}
  \mathbb{G}_{cur}\left(\tau_1,\tau_2\right)\left(s\right)=\left(cos(s)\right)\tau_1+\left(sin(s)\right)\frac{\tau_2-\tau_1cos(d({\tau_1,\tau_2}))}{sin(d({\tau_1,\tau_2}))},
  \label{eq:geo_line}
\end{equation}

\noindent where $d({\tau_1,\tau_2})=arccos(\tau_1 \bigodot \tau_2)$, indicating the Geodesic distance between $\tau_1$ and $\tau_2$. $\bigodot$ represents the dot product. The radian $s$, $0\le s\le d({\tau_1,\tau_2})$, controls the Geodesic distance between the newly generated data point and $\tau_1$. By incrementing $s$ progressively, a series of gradually changing Pre-Shapes can be generated.

Given two Pre-Shapes, we can generate new Pre-Shapes using Formula \ref{eq:geo_line}. However, only two Pre-Shapes constrain the representation of data distribution. We expect to employ additional Pre-Shapes to generate data points that better align with the actual distribution. In cases of multiple inputs, Formula \ref{eq:geo_line} becomes inapplicable. One type of approaches is to seek the optimal Geodesic curve within the Pre-Shape Space, minimizing the Geodesic distance to all input points \cite{han2010recognition,han2023fagcfeature}. The other type of approaches is to determine a Geodesic surface where all points on the surface constitute the generated Pre-Shapes. Generally, the Geodesic surface is defined in the tangent space \cite{friji2021geometric}, yet errors arise from projection into the tangent space and subsequent back-projection into the Pre-Shape Space. Pennec defines the Geodesic surface as Fréchet Barycentric Subspaces \cite{4f303e7f-6cd0-376a-9632-73d2741f0980}, which is hard to obtain data points on the Geodesic surface through the definition. We adopt the idea of some approaches that approximated the Geodesic surface with multiple Geodesic curves \cite{han2014recognizing,paskin2022kendall}.

\section{Methodology}
\label{sec:Methodology}

Our method, Information Transfer from the Built Geodesic surface (ITBGS), contains two module: Feature Augmentation on Geodesic Surface (FAGS); Interpolation and Regularization (I\&R).

\subsection{Feature Augmentation on Geodesic Surface (FAGS)}
\label{sec:Feature Augmentation on Geodesic Surface (FAGS)}
\begin{figure}[htbp]
\begin{center}
\includegraphics[width=1\linewidth]{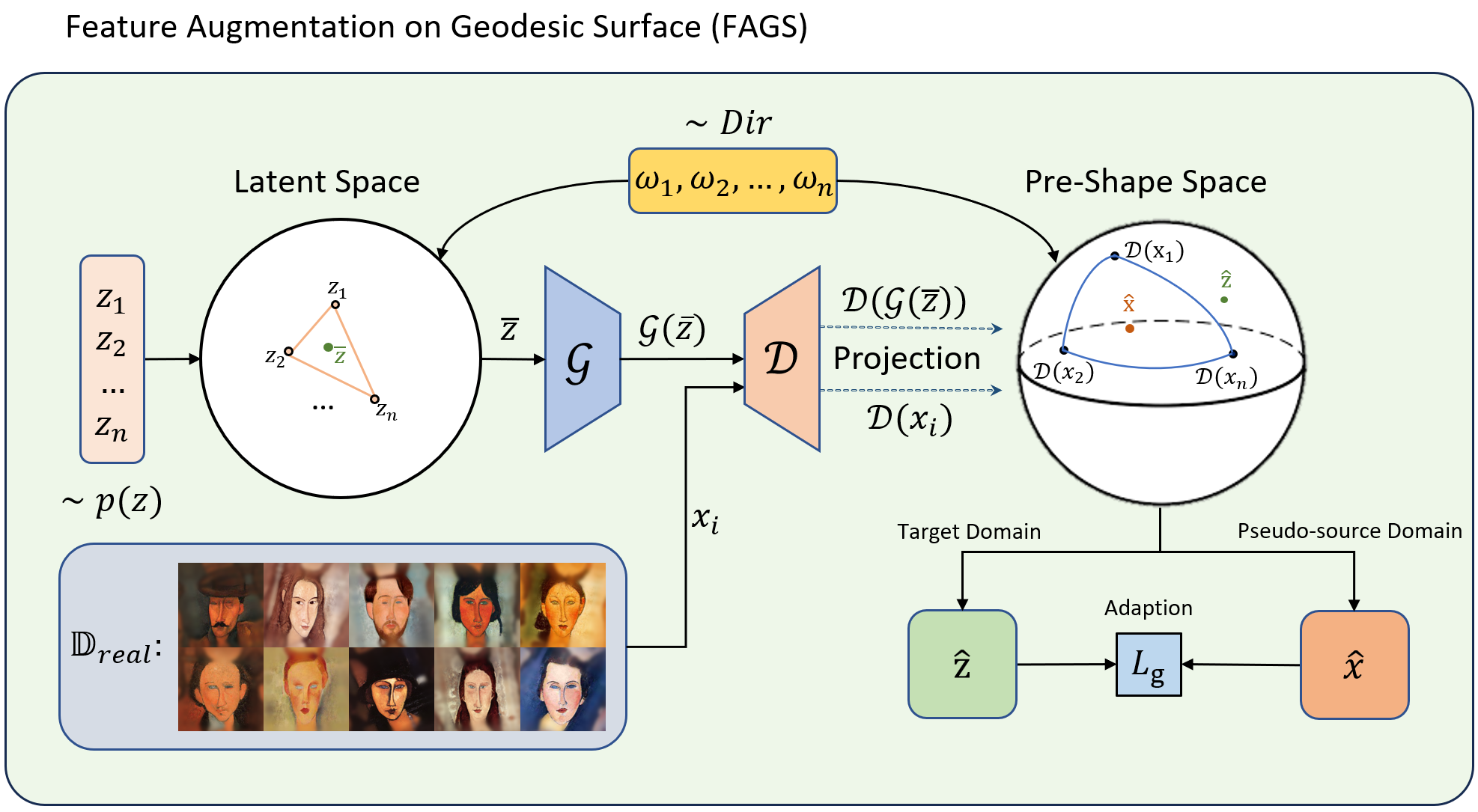}
\end{center}\caption{Illustration of the Feature Augmentation on Geodesic Surface (FAGS) module. We sample coefficients $\omega$ from the Dirichlet distribution and generate an anchor latent $\bar{z}$. Subsequently, we project the feature $\mathcal{D}(\mathcal{G}(\bar{z}))$ into the Pre-Shape Space. Similarly, we project the features extracted from the training set $\mathbb{D}_{real}$ denoted as $\mathcal{D}(x_i)$ and obtain new features $\hat{x}$ from the built Geodesic surface using the same weights $\omega$. Then, we ensure the self-correlation consistency between these two features, $\mathcal{D}(\mathcal{G}(\bar{z}))$ and $\hat{x}$.} 
\label{fig:FAGS}
\end{figure}


In fact, the information contained within the limited training samples remains underexploited and warrants further exploration. In light of this, we improve the original self-correlation consistency loss \cite{xiao_few_2022}, denoted as Geodesic self-correlation consistency loss, for the generator to capture structural information presented in the training samples. The more training samples we obtain, the richer and more accurate information they bring. Thus, for small samples, the data augmentation is especially valuable when the pre-trained models are absent. In line with existing practices, data augmentation has been widely adopted to bolster training datasets \cite{shorten2019survey}. We also utilize data augmentation to create a pseudo-source domain with small samples. Subsequently, we transfer the inherent structural information from the created pseudo-source domain to the target domain. We term this approach as Feature Augmentation on Geodesic Surface (FAGS). The illustration of FAGS is shown in Figure \ref{fig:FAGS}.

First, we create the pseudo-source domain by performing feature augmentation, as shown in Figure \ref{fig:motivation}. The features ought to be extracted from the samples to perform feature augmentation. Features extracted through GLCM \cite{haralick1973textural}, SIFT \cite{lowe1999object} and other deep learning based methods \cite{lecun2015deep} are all suitable for feature augmentation. We focus on generating images with small samples. Thus, GAN is adopted to obtain the image features. 

Most of recent studies have explored model inversion to deduce the features of input real images \cite{yin2020dreaming,tov2021designing,wang2022high}, obtaining intermediate feature maps from the generator $\mathcal{G}$. However, model inversion necessitates a trained generator $\mathcal{G}$. In every epoch, we input real images $x\sim \mathbb{D}_{real}$ into the $l$-th layer of the discriminator $\mathcal{D}$ to extract features $\mathcal{D}^l(x)$ during the training procedure. The extracted features are projected into the Pre-Shape Space to build the Geodesic surface. 

Generally, the Geodesic surface is defined in the tangent space \cite{friji2021geometric}, yet errors arise from projection into the tangent space and subsequent back-projection into the Pre-Shape Space. Han et al. \cite{han2014recognizing} approximated a Geodesic surface with multiple Geodesic curves. A Geodesic surface $\mathbb{G}_{FBS}(\tau,\omega)$ in the Pre-Shape Space can be noted as Fréchet Barycentric Subspaces (FBS) \cite{4f303e7f-6cd0-376a-9632-73d2741f0980}, denoted as follows:

\begin{equation}
  \mathbb{G}_{FBS}(\tau ,\omega)=\left\{ \arg\min_\mu{\sum_{j=1}^{n}\omega_jd(\mu,\tau_j):\ \sum_{j=1}^{n}\omega_j\neq0} \right\}.
  \label{eq:FBS}
\end{equation}

$\mu$ is a vector and a Pre-Shape on the Geodesic surface $\mathbb{G}_{FBS}(\tau ,\omega)$. $\tau\triangleq \{\tau_1, ...,\tau_n\}$ and $\omega\triangleq\{\omega_1,... ,\omega_n\}$, representing two sets of the given vectors and weights, respectively. $n$ denotes the number of the given vectors. Since Formula \ref{eq:FBS} is difficult to calculate, we adopt a equivalent way to accomplish the calculation with iteratively building the Geodesic curves $\mathbb{G}_{cur}(\cdot)$, from the paper \cite{paskin2022kendall}. Thus, Formula \ref{eq:FBS} can be rewritten to Formula \ref{eq:geo_surface} based on Formula \ref{eq:geo_line}, as follows:

\begin{equation} \label{eq:geo_surface}
  \mu_j=\mathbb{G}_{cur}(\mu_{j-1}, \tau_j)(\frac{\omega_j}{\sum^j_{i=1}\omega_i}),\ {\rm where}\ j=2,...,n,\ 
\end{equation}

\noindent where $\mu_1=\tau_1$. Thus, when $j=n$, the Geodesic surface $\mathbb{G}_{surf}(\cdot)$ can be built with a set of vectors $\tau$ and a set of weights $\omega$. The formula is defined as follows:

\begin{equation}
  \mathbb{G}_{surf}(\tau,\omega)=\mu_n.
  \label{eq:g_surf}
\end{equation}

$\mathcal{D}^l(x)\in\mathbb{R}^{c\times h\times w}$ denotes the features extracted from the $l$-th layer of the discriminator $\mathcal{D}$. $\mathcal{R}(\mathcal{D}^l(x)) \in\mathbb{R}^{2 \times (chw/2)}$ reshapes the dimension of feature vectors $\mathcal{D}^l(x)$ from $c \times h \times w$ to $2 \times (chw/2)$, representing a set of $chw/2$ points. Each point indicates coordinates in a 2D space, so that $\mathcal{R}(\mathcal{D}^l(x))$ can be easily projected into the Pre-Shape Space. We define the entire projection function into the Pre-Shape Space as $f_p(\cdot)=\mathcal{V}(\mathcal{Q}(\mathcal{R}(\cdot)))$, based on Formula \ref{eq:mean-reduction} and \ref{eq:normalization}. Thus, from the built Geodesic surface $\mathbb{G}_{surf}(f_p(\mathcal{D}^l(x)),\omega)$, multiple new feature vectors $\hat{x}^l$ can be calculated and constitute of a pseudo-source domain, denoted as $D_{ps}$.


Correspondingly, we preprocess a target domain, as shown in Figure \ref{fig:motivation}. the anchor latent $\bar{z}$ \cite{kong_few-shot_2022} is calculated in the target domain using the same weights $\omega$, as follows:

\begin{equation}
  \bar{z}=\sum^n_{i=1}\omega_iz_i,
  \label{eq:latent}
\end{equation}

\noindent where $\{z_i:i\in[1,n]\}$ denotes a set of random latent vectors. The anchor image $\mathcal{G}(\bar{z})$ can be obtained by inputting $\bar{z}$ into generator $\mathcal{G}$. To extract the features from $\mathcal{G}(\bar{z})$ for the target domain, we input $\mathcal{G}(\bar{z})$ into discriminator $\mathcal{D}$. Thus, multiple feature vectors $\hat{z}^l$ from the $l$-th layer of $\mathcal{D}$ constitute of the target domain $D_t$, where $\hat{z}^l=f_p(\mathcal{D}^l(\mathcal{G}(\bar{z})))$. 

We expect to transfer information from the pseudo-source domain $D_{ps}$ to the target domain $D_t$. The Geodesic self-correlation consistency loss $L_g$ serves to enforce the inherent structural relationships between features of $D_{ps}$ and $D_t$.

First, we reshape the dimension of $\hat{x}^l$ and $\hat{z}^l$ back to $c \times h \times w$. Let $\hat{x}^l(u,v)$ and $\hat{z}^l(u,v)$ signify vectors with $c$ dimensions located at the position $(u,v)$ of $\hat{x}^l$ and $\hat{z}^l$, respectively. The cosine similarity $C^{\hat{x}^l}_{u,v}(a,b)$ of $\hat{x}^l$ between position $(u,v)$ and its corresponding position $(a,b)$ can be calculated as follows:


\begin{equation}
  C^{\hat{x}^l}_{u,v}(a,b)=\frac{<\hat{x}^l(u,v), \hat{x}^l(a,b)>}{\|\hat{x}^l(u,v)\|\cdot\|\hat{x}^l(a,b)\|}.
  \label{eq:selfcorr_src}
\end{equation}

After traversing all spatial corresponding positions, we obtain a self-correlation matrix $C^{\hat{x}^l}_{u,v}$. Similarly, we can calculate the cosine similarity $C^{\hat{z}^l}_{u,v}(a,b)$ of $\hat{z}^l$ between position $(u,v)$ and its corresponding position $(a,b)$, as follows:

\begin{equation}
  C^{\hat{z}^l}_{u,v}(a,b)=\frac{<\hat{z}^l(u,v), \hat{z}^l(a,b)>}{\|\hat{z}^l(u,v)\|\cdot\|\hat{z}^l(a,b)\|},
  \label{eq:selfcorr_tar}
\end{equation}

\noindent and obtain a self-correlation matrix $C^{\hat{z}^l}_{u,v}$. As a result, $L_g$ can be formulated as follows:

\begin{equation}
  L_g=\mathbb{E}_{z\sim p(z),x\sim \mathbb{D}_{real}, \omega\sim Dir}\sum_l\sum_{u,v}L_{s\ell1}(C^{\hat{x}^l}_{u,v}, C^{\hat{z}^l}_{u,v}),
  \label{eq:lgscc}
\end{equation}

\noindent where $l$ iterates over the selected convolutional layers of the feature extractor. $(u,v)$ traverses all spatial positions and $L_{s\ell1}(\cdot)$ represents the smooth-$\ell$1 loss function \cite{ren2015faster}.

\subsection{Interpolation and Regularization (I$\&$R)}
\label{sec:Interpolation Supervision and Regularization (InSR)}

\begin{figure}[htbp]
  \begin{center}
  \includegraphics[width=1\linewidth]{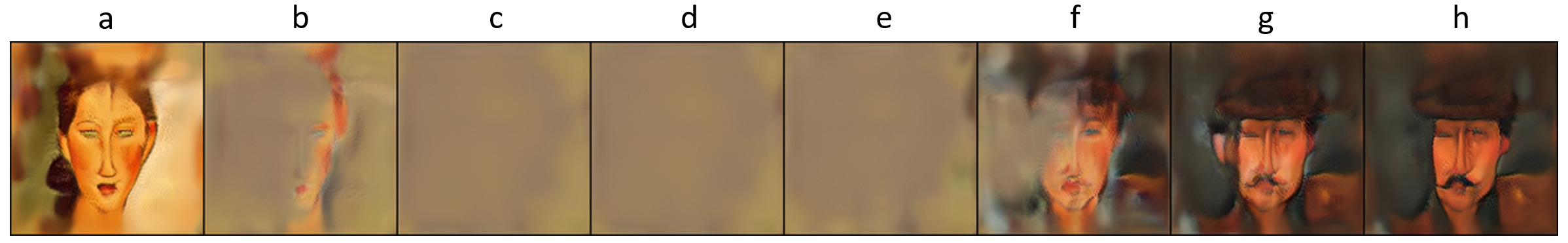}
  \end{center}\caption{Interpolated images generated by StyleGAN2 with FAGS. Blurriness occurs in the intermediate interpolations.} 
  \label{fig:interp}
\end{figure}

The importance of observing the latent space interpolation is to mitigate the potential occurrence of blurry and "stairlike" interpolation outcomes. Smooth latent space interpolation is an important property of generative models that disproves overfitting and allows synthesis of novel data samples \cite{kong_few-shot_2022}.
The interpolation set in the latent space $\{z'_1,z'_2,...,z'_k\}$, predefined as $Z_{inp}(z'_1,z'_k)$, which is obtained by linear interpolating two random latent vectors $z'_1,z'_k \sim p(z)$. Subsequently, we can obtain a set of interpolated images through $\mathcal{G}(Z_{inp})$ with FAGS employed during the training of $\mathcal{G}$. However, the changes in the appearances of $\mathcal{G}(Z_{inp})$ are discontinuous, causing the occurrence of blurriness in the intermediate interpolations. For instance, Figure \ref{fig:interp} visually portrays one of the $\mathcal{G}(Z_{inp})$ through the trained StyleGAN2 \cite{karras_analyzing_2020} with FAGS.



To alleviate the blurry appearance and generate more realistic images, we present an interpolation strategy. To supervise $\mathcal{G}(Z_{inp})$, the original adversarial loss \cite{goodfellow_generative_2020} can be rewriten based on Formula \ref{eq:Lg_adv} and \ref{eq:Ld_adv} for both generator $\mathcal{G}$ and discriminator $\mathcal{D}$, as follows:


\begin{equation}
  L_{inp}=\mathbb{E}_{z'_1,z'_k\sim p(z)}[log(\mathcal{D}(\mathcal{G}(Z_{inp}(z'_1,z'_k))))].
  \label{eq:L_interp}
\end{equation}


\begin{figure}[htbp]
\begin{center}
\includegraphics[width=1\linewidth]{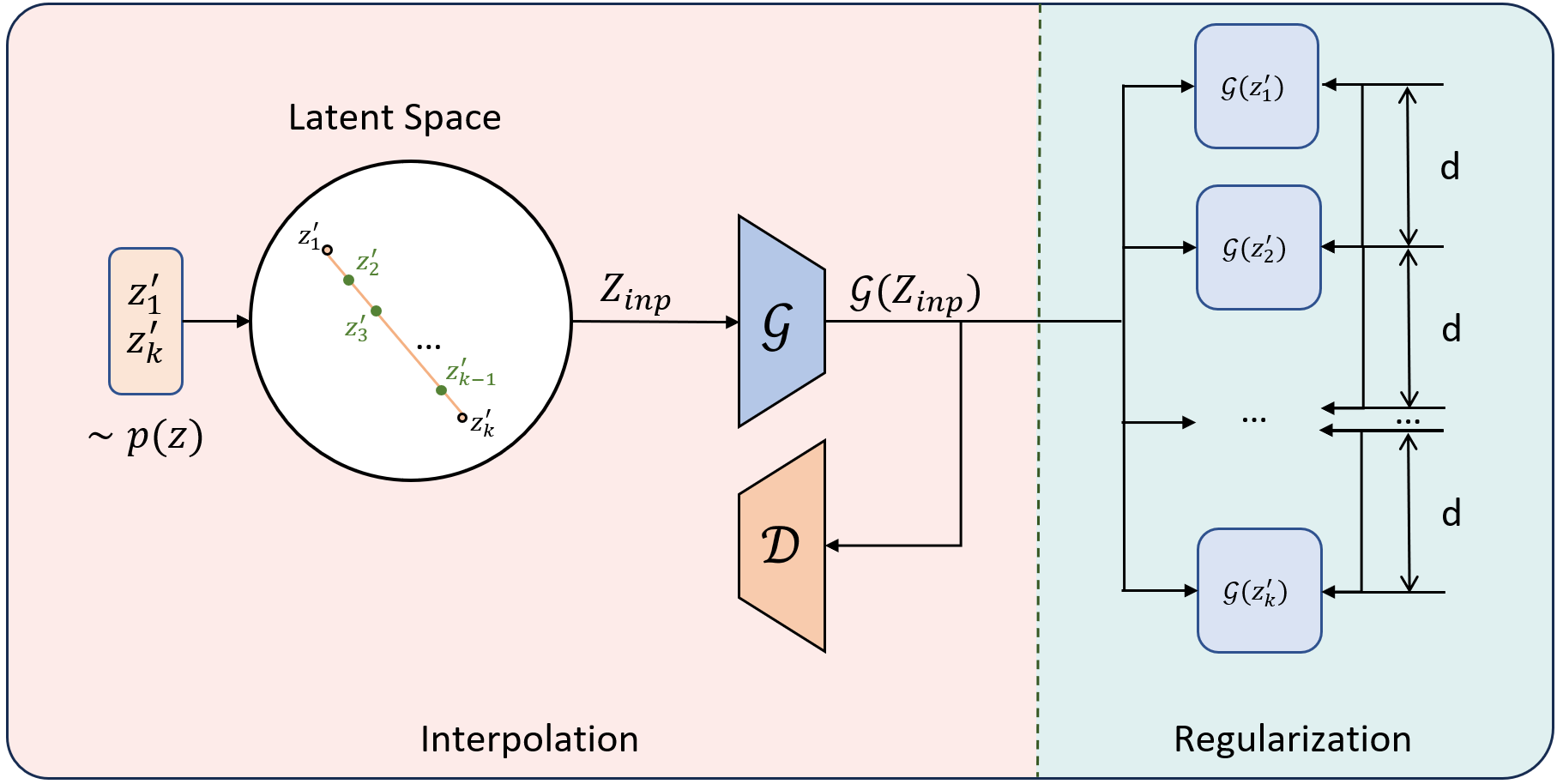}
\end{center}\caption{Illustration of the I\&R module, containing two parts: Interpolation and Regularization.} 
\label{fig:InSR}
\end{figure}

To further avoid ``stairlike" phenomenon, we propose a distance regularization strategy. We adopt the KL-Divergence loss \cite{kullback1951information} to enforce equal pairwise distances between features of interpolated images, namely $L_{dr}$. Average pooling is used to aggregate the feature map for minimizing the computational time and space, reducing both the $H$ and $W$ down to 1/4 of the original size. The proposed strategy penalizes the generator $\mathcal{G}$ from merely memorizing real samples and helps to generate smooth images during latent space interpolation. 

The two strategies consist of a module, named as Interpolation and Regularization (I\&R), as shown in Figure \ref{fig:InSR}. Algorithm \ref {alg:regualrization} outlines the pseudo-code of the I\&R module in a PyTorch-like style.

\begin{algorithm}
\caption{Pseudo-code of I\&R module.}\label{alg:regualrization}
\begin{algorithmic}[1]
  \small
\Require $\rm{z1,zk}: random\ latents$
\Require $\rm k: size\ of\ interpolations$
\State $\rm \#\ \  Interpolation $
\State $\rm Z\_inp = cat([lerp(z1, zk, v)\ for\ v\ in\ linspace(0, 1, k)])$
\State $\rm inp\_imgs, inp\_feats = Generator(Z\_inp)$
\State $\#\ inp\_feats:k \times c \times h \times w$
\State $\rm pred = Discriminator(inp\_imgs)$
\State $\rm L\_inp = log(pred\_imgs).mean()$

\State
\State $\rm \#\ \  Regularization $
\State $\rm dist() = L2\_distance()$
\State $\rm inp\_feats = AdaptiveAvgPool2d(inp\_feats)$
\State $\#\ k \times c \times h \times w \rightarrow k \times c \times h/4 \times w/4$
\State $\rm inp\_feats\_temp = cat([inp\_feats[1:], inp\_feats[0]])$
\State $\rm feats\_dist = dist(inp\_feats, inp\_feats\_temp)$
\State $\rm q\_dist = cat([ones(k-1), Tensor([k-1])])$
\State $\rm L\_dr = KLDivLoss(feats\_dist, q\_dist)$
\end{algorithmic}
\end{algorithm}

\subsection{Final Optimization Function}
\label{sec:Final Optimization Function}


The final optimization function $L^\mathcal{G}$ for generator $\mathcal{G}$ and $L^\mathcal{D}$ for discriminator $\mathcal{D}$ are defined as follows:
\begin{equation} \label{eq:Lg}
  L^\mathcal{G}=L^\mathcal{G}_{adv}-\lambda_1L_{inp}+\lambda_2L_{dr},
\end{equation}
\noindent and
\begin{equation} \label{eq:Ld}
  L^\mathcal{D}=L^\mathcal{D}_{adv}+\lambda_1L_{inp}+\lambda_3L_g,
\end{equation}

\noindent where $L_{dr}$ and $L_g$ are only applied to update $\mathcal{G}$ and $\mathcal{D}$, respectively. $L_{inp}$ is employed on updating both $\mathcal{G}$ and $\mathcal{D}$. $\lambda_1, \lambda_2$ and $\lambda_3$ are fixed ratio parameters.

\section{Experiments and Discussion}
\label{sec:Experiments and Discussion}
\subsection{Implementation details}
\label{sec:Implementation details}

We employ StyleGAN2 \cite{karras_analyzing_2020} with MixDL \cite{kong_few-shot_2022} as our backbone architecture without using ADA \cite{karras2020training} to deal with the extreme few-shot scenario. We set the parameters of Formula \ref{eq:Lg} and \ref{eq:Ld} as follows: $\lambda_1=0.8, \lambda_2=1.25,$ and $\lambda_3=0.8$. Both batch size and interpolation size are configured as 4 to ensure compatibility with a single Nvidia GeForce RTX 3090 (24GB) for training.

Our experiments involve qualitative and quantitative comparisons with several models, including N-div \cite{liu2019normalized}, MSGAN (MG)  \cite{mao2019mode}, DistanceGAN (DG) \cite{benaim2017one}, StyleGAN2 (SG2) \cite{karras_analyzing_2020}, StyleGAN2+ADA (SG2A) \cite{karras2020training}, FastGAN (FG) \cite{liu_towards_2021}, and MixDL (MDL) \cite{kong_few-shot_2022}.

We experiment on multiple datasets, including Amedeo Modigliani paintings \cite{yaniv2019face}, Landscape drawings \cite{ojha_few-shot_2021}, Animal-Face Dog \cite{si2011learning}, Face sketches \cite{wang2008face}, Anime face \cite{liu_towards_2021}, Pokemon \cite{liu_towards_2021}, FFHQ \cite{karras_style-based_2019}, and CelebA \cite{liu2018large}. Specifically, Amedeo Modigliani paintings and Landscape drawings are 10-shot datasets. For the remaining datasets, which contain more than 10 images, 10 images are randomly chosen from each to construct the 10-shot subsets. All images in our experiments are of resolution 256 $\times$ 256.

\subsection{Qualitative Comparison}
\label{sec: Qualitative Comparison}

\begin{figure*}[htbp]
  \begin{center}
  \includegraphics[width=1\linewidth]{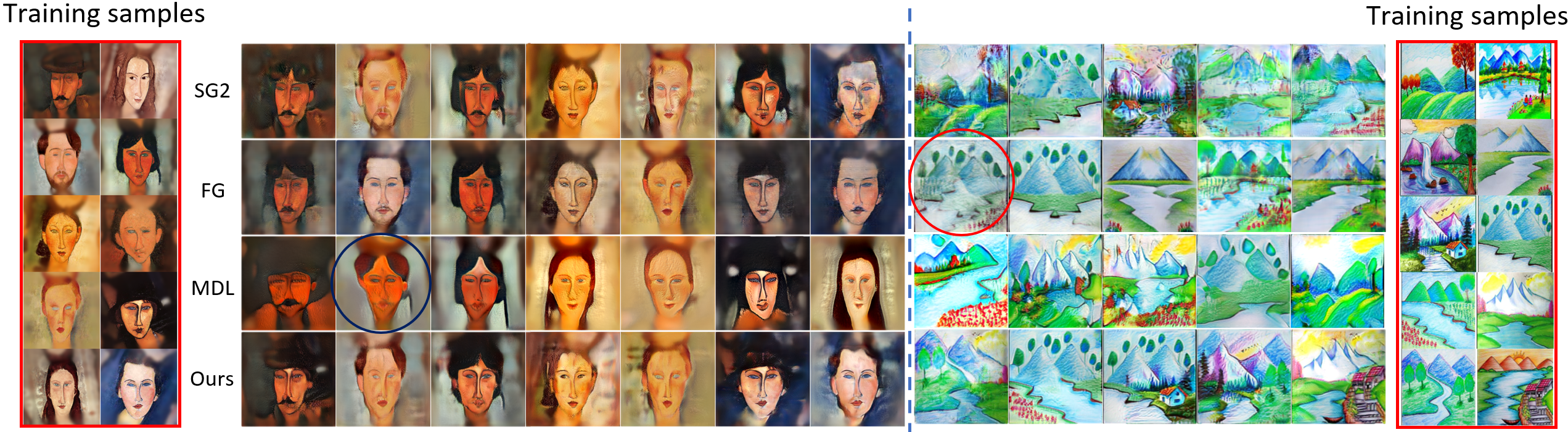}
  \end{center}\caption{Training and generated samples of several methods on Amedeo Modigliani paintings (left)  and Landscape drawings (right). Our method exhibits more fidelity and diversity.} 
  \label{fig:ame_land_quali}
  \end{figure*}

In Figure \ref{fig:ame_land_quali}, the results of various methods are presented on the Amedeo Modigliani paintings and Landscape drawings datasets. It is important to note that all the models are trained from scratch without auxiliary information produced, except for FastGAN \cite{liu_towards_2021}, which utilized a pre-trained VGG for calculating perceptual loss \cite{zhang_unreasonable_2018}. From Figure \ref{fig:ame_land_quali}, the generated samples of StyleGAN2 \cite{karras_analyzing_2020} show blurriness and overfitting on both 10-shot datasets. FastGAN \cite{liu_towards_2021} produces comparable results to our method on the Amedeo Modigliani paintings dataset, but demonstrates inferior quality on the Landscape drawings dataset. FastGAN merely weighted add two training samples, as depicted in the generated sample circled in red in Figure \ref{fig:ame_land_quali}. MixDL-generated images \cite{kong_few-shot_2022} excel in terms of diversity but lack in fidelity compared to other methods. For example, the face shape is distorted in the generated sample circled in blue in Figure \ref{fig:ame_land_quali}. Notably, ITBGS stands out by generating images with both sufficient fidelity and diversity on both 10-shot datasets. The achievement is attributed to the natural fusion of visual elements, such as shapes, colors, textures, from two or more real images. We hypothesize that the features generated on the Geodesic surface, constructed in the Pre-Shape Space using extracted image features from the dataset, represent a natural integration of from multiple image features. Consequently, these generated features facilitate the generation of images that exhibit a more natural fusion of visual elements.

\begin{figure}[htbp]
\begin{center}
\includegraphics[width=1\linewidth]{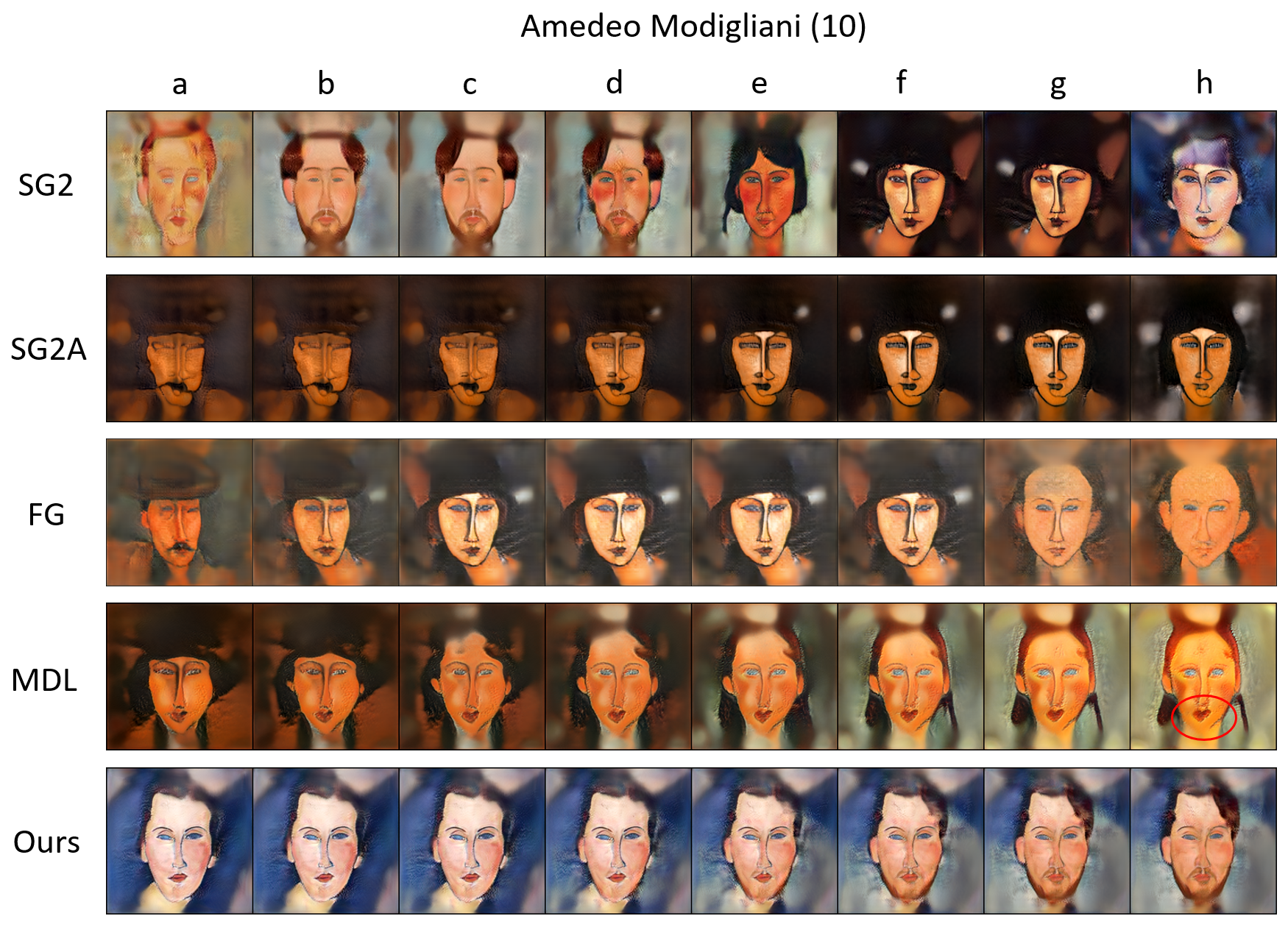}
\end{center}\caption{Latent space interpolation results on Amedeo Modigliani paintings.} 
\label{fig:ame_interp_quali}
\end{figure}

\begin{figure}[htbp]
  \begin{center}
  \includegraphics[width=1\linewidth]{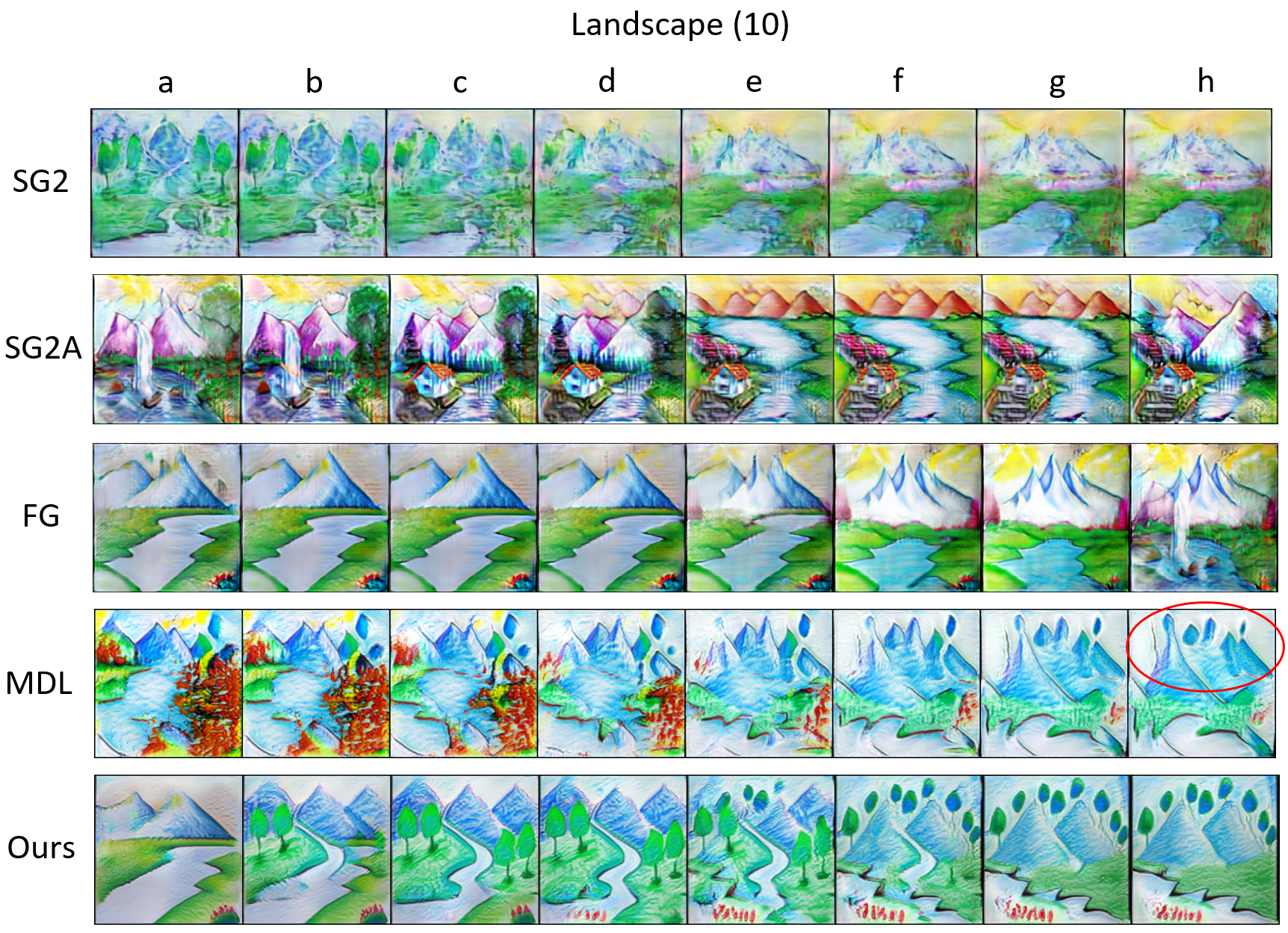}
  \end{center}\caption{Latent space interpolation results on Landscape drawings.} 
  \label{fig:land_interp_quali}
  \end{figure}

\begin{figure}[htbp]
\begin{center}
\includegraphics[width=1\linewidth]{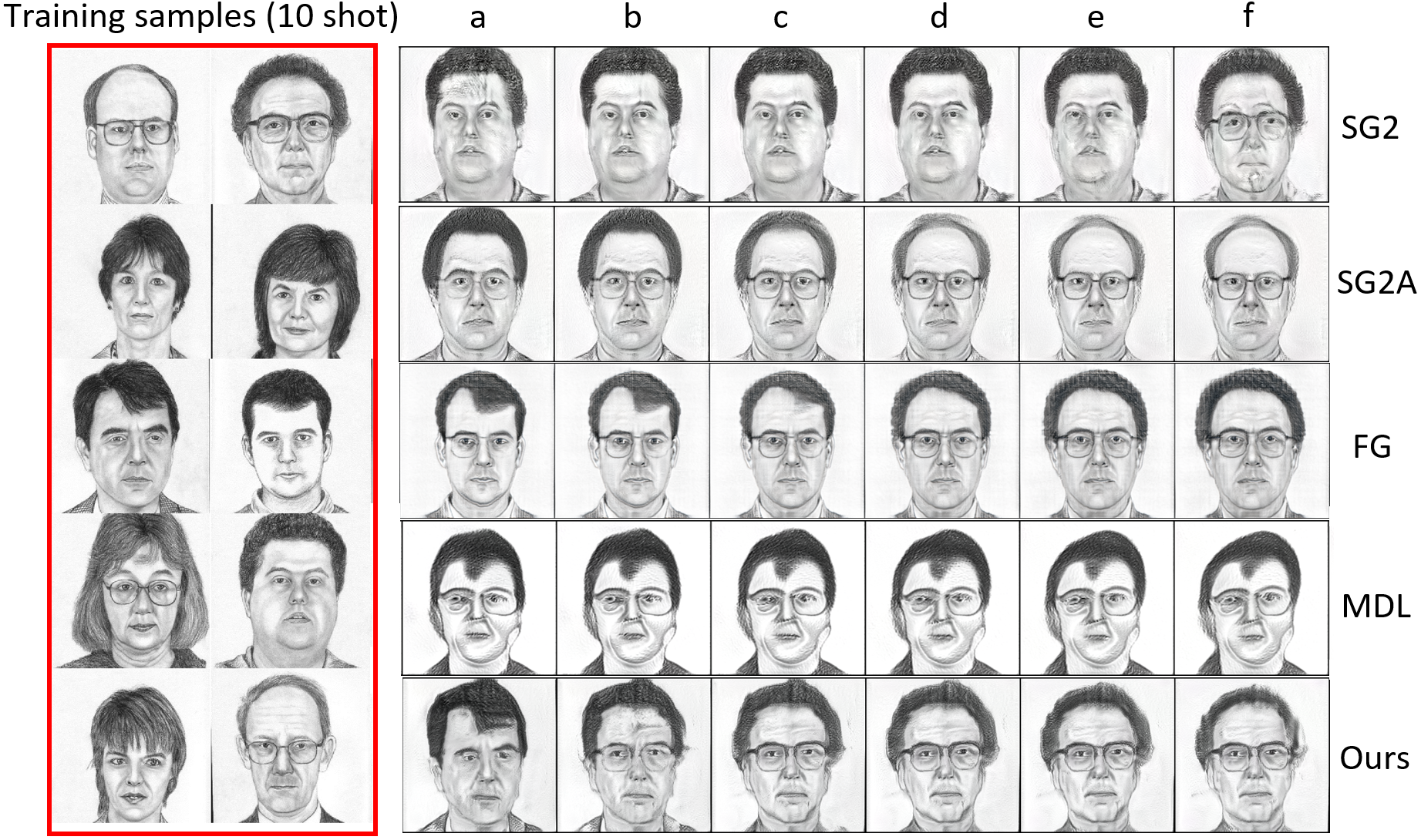}
\end{center}\caption{Latent space interpolation results on Face Sketches.} 
\label{fig:sketches_interp_quali}
\end{figure}

\begin{figure}[htbp]
  \begin{center}
  \includegraphics[width=1\linewidth]{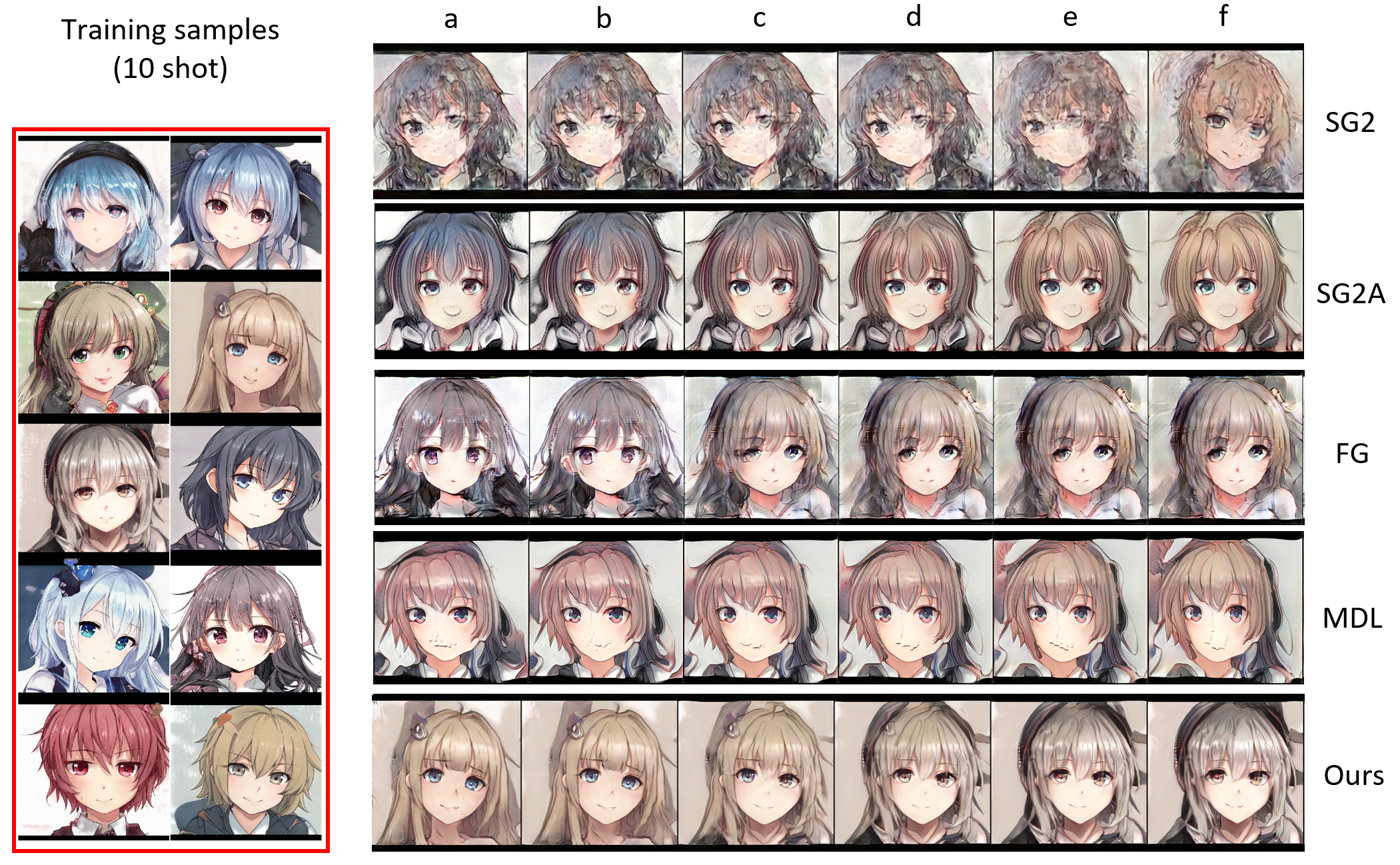}
  \end{center}\caption{Latent space interpolation results on Anime Face.} 
  \label{fig:anime_interp_quali}
  \end{figure}



In Figure \ref{fig:ame_interp_quali}, \ref{fig:land_interp_quali}, \ref{fig:sketches_interp_quali} and \ref{fig:anime_interp_quali}, we provide a comparison of the interpolated samples generated with some methods across various 10-shot datasets. Notably, FastGAN \cite{liu_towards_2021} achieves comparable generative results to our method on the Face sketches dataset. However, FastGAN-generated interpolated samples exhibit the ``stairlike" phenomenon from column f to column g in Figure \ref{fig:ame_interp_quali} (Amedeo Modigliani). Also, defects occur in column c of Figure \ref{fig:anime_interp_quali}. MixDL \cite{kong_few-shot_2022} excels in producing smooth semantic interpolations but comes at the cost of reduced fidelity. For instance, the mouth and the mountain peaks appear strange in the red-circled sections of Figure \ref{fig:ame_interp_quali} and \ref{fig:land_interp_quali}, respectively. StyleGAN2+ADA \cite{karras2020training} also demonstrates a similar trade-off of fidelity and diversity on Amedeo Modigliani and Anime Face dataset. In contrast, the original StyleGAN2 \cite{karras_analyzing_2020} generates interpolated images with acceptable fidelity but exhibits the ``stairlike" phenomenon, as depicted from column e to column f in Figure \ref{fig:ame_interp_quali} and \ref{fig:sketches_interp_quali}. ITBGS shows the capability of achieving smooth latent space interpolation across all 10-shot datasets while maintaining sufficient fidelity. Smooth latent space interpolation underlines the effectiveness of ITBGS in balancing fidelity and diversity for image generation.

\begin{figure}[htbp]
\begin{center}
\includegraphics[width=1\linewidth]{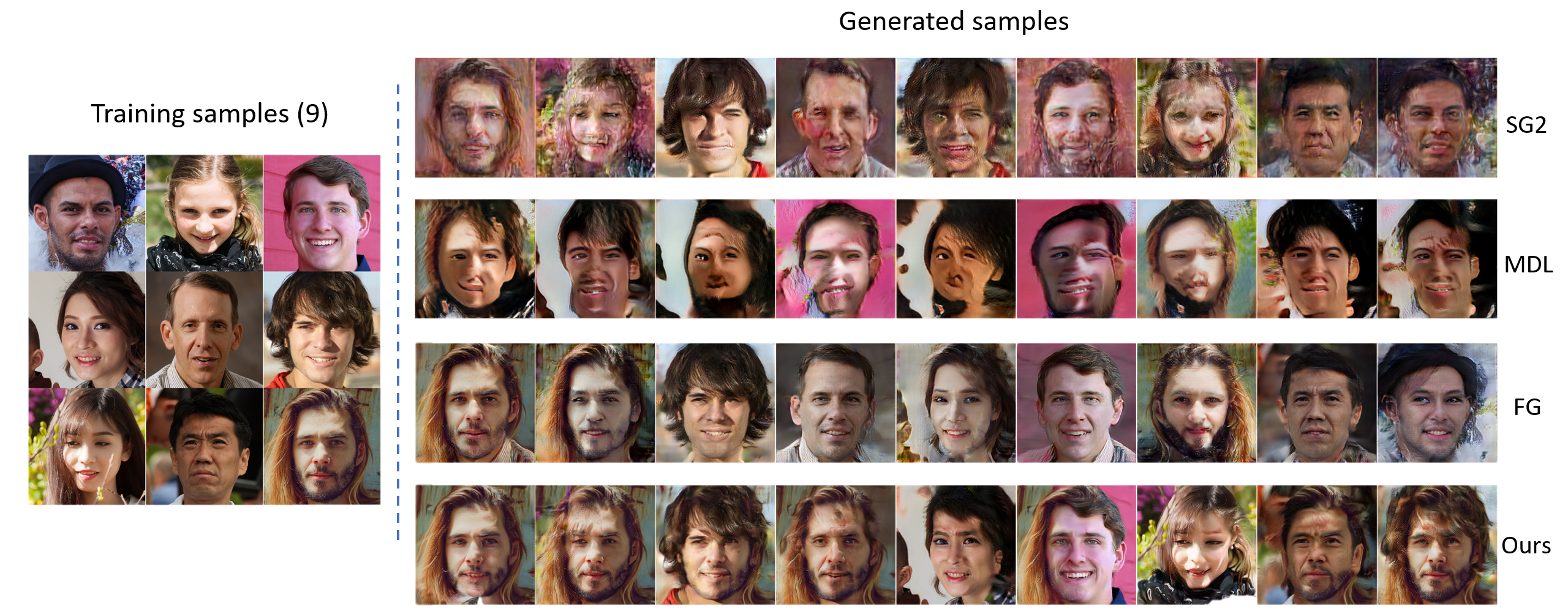}
\end{center}\caption{10-shot image generation results on FFHQ subset.} 
\label{fig:ffhq_quali}
\end{figure}

The quality of results in generating real-world images serves as a valuable metric for evaluating the effectiveness of generative models. The experiments on real-world face datasets FFHQ \cite{karras_style-based_2019}, as depicted in Figure \ref{fig:ffhq_quali}, demonstrate a relatively satisfactory ability to combine facial features from two or more faces. The ability of combination is particularly evident in the seamless blending of hairstyles, beards, and other facial attributes. Notably, FastGAN \cite{liu_towards_2021} also produces decent results on the FFHQ dataset, while the other comparative methods struggle to generate real-world facial images.

\subsection{Quantitative comparison}
\label{sec:Quantitative comparison}

\begin{table*}[htbp]
  \centering
  \caption{Quantitative results on 10-shot image generation task. The best and the second best scores are in bold and underlined, respectively.} 
  \label{table:all_quanti}
    \resizebox{\textwidth}{25mm}{
    \setlength{\tabcolsep}{0.3mm}{
  \begin{tabular}{@{}ccccccccccccc@{}}
  \toprule
  \multirow{2}{*}{Method}                                   & \multicolumn{2}{c}{Anime-Face} & \multicolumn{2}{c}{Animal Dog} & \multicolumn{2}{c}{Face Sketches} & \multicolumn{2}{c}{Amedeo Modigliani} & \multicolumn{2}{c}{Landscapes} & \multicolumn{2}{c}{Pokemon}     \\ \cmidrule(l){2-13} 
                                                            & FID(↓)        & LPIPS(↑)       & FID(↓)        & LPIPS(↑)       & FID(↓)          & LPIPS(↑)        & FID(↓)            & LPIPS(↑)          & FID(↓)          & LPIPS(↑)         & FID(↓)         & LPIPS(↑)       \\ \midrule
  N-Div\cite{liu2019normalized}           & 175.4         & 0.425          & 150.4         & 0.632          & /               & /               & /                 & /                 & /               & /                & /              & /              \\
  MSGAN\cite{mao2019mode}                 & 138.6         & 0.536          & 165.7         & 0.630          & /               & /               & /                 & /                 & /               & /                & /              & /              \\
  DistanceGAN\cite{benaim2017one}         & \uline{84.1}          & \textbf{0.543} & 102.6         & 0.678          & /               & /               & /                 & /                 & /               & /                & /              & /              \\
  StyleGAN2\cite{karras_analyzing_2020} & 213.9         & 0.407          & 312.9         & 0.549          & 188.4           & \uline{0.476}     & \textbf{68.6}     & \textbf{0.649}    & 210.3           & 0.531            & 261.9          & 0.475          \\
  StyleGAN2+ADA\cite{karras2020training}  & 282.3         & 0.473          & 342.0         & 0.539          & 341.3           & 0.469           & 216.3             & 0.538             & 207.7           & 0.498            & 278.5          & 0.413          \\
  MixDL\cite{kong_few-shot_2022}        & 140.9         & 0.529          & 291.1         & \uline{0.701}    & 137.9           & 0.396           & 205.2             & 0.643             & 183.3           & \uline{0.698}   & 231.2          & 0.499          \\
  FastGAN\cite{liu_towards_2021}        & 150.5         & 0.393          & \uline{65.1} & 0.671          & 112.4           & 0.437           & 108.3       & 0.615             & \uline{ 83.8}   & 0.689      & \uline{203.3} & \textbf{0.554} \\ \midrule
  FastGAN+FAGS                         & 123.2          & 0.304          & \textbf{54.5}          & 0.679          & \uline{97.8}            & 0.292           & \uline{98.9}              & 0.588             & \textbf{82.6}              & \textbf{0.699}           & \textbf{200.8}            & 0.420         \\
  Ours                                                      & \textbf{72.5} & \uline{0.538}    & 95.0    & \textbf{0.713} & \textbf{57.7}   & \textbf{0.485}  & 113.9             & \uline{0.647}       & 90.7      & 0.677            & 208.1    & \uline{0.552}    \\ \bottomrule
  \end{tabular}}}
  \end{table*}

In the evaluation, we employ the Fréchet Inception Distance (FID) \cite{heusel2017gans} as well as the pairwise Learned Perceptual Image Patch Similarity (LPIPS) \cite{zhang_unreasonable_2018} as metrics. FID is computed against the few-shot dataset, while LPIPS is calculated between generated samples. Lower FID values and higher LPIPS values are indicative of higher image quality and diversity, respectively.

\begin{table}[htbp]
  \centering
  \caption{Quantitative results on FFHQ and CelebA. The best and the second best scores are in bold and underlined, respectively.} 
  \begin{tabular}{@{}ccccc@{}}
    \toprule
    \multicolumn{1}{l}{\multirow{2}{*}{Method}} & \multicolumn{2}{c}{FFHQ}        & \multicolumn{2}{c}{CelebA}     \\ \cmidrule(l){2-5} 
    \multicolumn{1}{l}{}                        & FID(↓)         & LPIPS(↑)       & FID(↓)        & LPIPS(↑)       \\ \midrule
    StyleGAN2                                   & 311.6          & 0.442          & 102.3         & \uline{0.561}    \\
    MixDL                                       & 283.7          & \textbf{0.640} & 206.8         & 0.531          \\
    FastGAN                                     & \textbf{112.0} & 0.593          & \uline{86.6}    & 0.507          \\ \midrule
    FastGAN+FAGS                                & 220.9          & 0.448              & \textbf{67.3}          & 0.554 \\
    Ours                             & \uline{130.9}    & \uline{0.617}    & 91.3   & \textbf{0.570} \\ \bottomrule
  \end{tabular}
  \label{table:ffhq_quanti}
  \end{table}

  Tables \ref{table:all_quanti} and \ref{table:ffhq_quanti} present the quantitative results of the compared methods on various 10-shot datasets. We use consistent parameter settings across different datasets and conduct no domain-specific fine-tuning. The tables show that our method achieves the best or near-best results in terms of both FID and LPIPS. Assessing the capabilities of generative model requires considering both fidelity and diversity, which are reflected by these metrics. While the performance metrics of our method are slightly lower than FastGAN \cite{liu_towards_2021} on Pokemon and Landscape datasets, it still ranks the third-best position. On the real-world face dataset, the generated samples by our method and FastGAN yield comparable visual quality, as demonstrated in Figure \ref{fig:ffhq_quali}. However, when examining the quantitative metrics in Table. \ref{table:ffhq_quanti}, our method outperforms FastGAN. While FastGAN has a slightly lower FID, our method excels in terms of diversity. Furthermore, the integration of the FAGS module into FastGAN results in enhanced performance on some of the datasets, including Pokemon, Landscapes, and CelebA datasets.

\subsection{Ablation Study}
\label{sec:Ablation Study}

\begin{table}[htbp]
\centering
\caption{Quantitative ablation on the proposed modules.}
\begin{tabular}{@{}cccccc@{}}
\toprule
\multirow{2}{*}{Backbone} & \multirow{2}{*}{FAGS} & \multirow{2}{*}{I} & \multirow{2}{*}{R} & \multicolumn{2}{c}{Landscapes} \\ \cmidrule(l){5-6} 
                          &                       &                                &                                 & FID(↓)        & LPIPS(↑)       \\ \midrule
\multirow{4}{*}{StyleGAN2} & ×                     & ×                              & ×                               & 210.3         & 0.531          \\
                          & \checkmark                     & ×                              & ×                               & 206.9(-3.4)   & 0.629(+0.098)  \\
                          & \checkmark                     & \checkmark                              & ×                               & 182.3(-24.6)  & 0.669(+0.04)   \\
                          & \checkmark                     & \checkmark                              & \checkmark                               & \textbf{90.7(-91.6)}   & \textbf{0.677(+0.008)}  \\ \midrule
\multirow{3}{*}{FastGAN}  & ×                     & ×                              & ×                               & 83.8          & 0.689          \\
                          & \checkmark                     & ×                              & ×                               & \textbf{82.6(-1.2)}    & \textbf{0.699(+0.01)}   \\
                          & \checkmark                     & \checkmark                              & \checkmark                               & 97.5(+14.9)   & 0.679(-0.02)   \\ \bottomrule
\end{tabular}
\label{table:ablation_all}
\end{table}

We conduct quantitative ablation experiments with the proposed modules. As shown in Table \ref{table:ablation_all}, the integration of the FAGS module enhances the performance metrics for both backbones: StyleGAN2 and FastGAN. The deployment of the I\&R module contributes to a further performance increment for StyleGAN2, since adding only FAGS may results in blurry generated images, as shown in Figure \ref{fig:interp}. However, for FastGAN, adding the I\&R module results in a inferior performance compared to the sole addition of the FAGS module, which suggests that the I\&R module may be backbone-sensitive.

\begin{figure}[htbp]
  \begin{center}
  \includegraphics[width=1\linewidth]{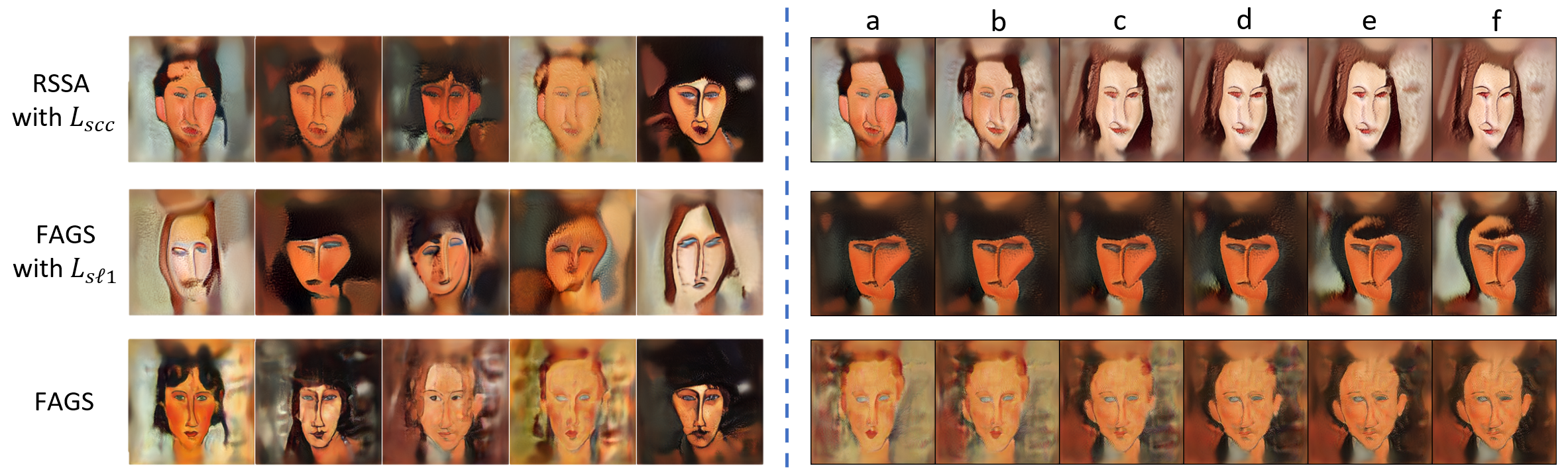}
  \end{center}\caption{Qualitative ablation on FAGS module.} 
  \label{fig:ame_ablation_quali}
  \end{figure}

  \begin{table}[htbp]
    \centering
    \caption{Quantitative ablation on FAGS module.}
    \begin{tabular}{@{}ccc@{}}
    \toprule
    \multirow{2}{*}{Method} & \multicolumn{2}{c}{Amedeo Modigliani} \\ \cmidrule(l){2-3} 
                                     & FID(↓)                 & LPIPS(↑)              \\ \midrule
    RSSA with $L_{scc}$                            & 186.8                & 0.585                \\
    FAGS with $L_{s\ell1}$                   & 187.1              & 0.527                \\
    FAGS with $L_g$                  & \textbf{113.9}       & \textbf{0.647}       \\ \bottomrule
    \end{tabular}
    \label{table:ablation}
    \end{table}

\subsubsection{Effect of the FAGS module}
\label{sec:Effect of the FAGS}
To validate the effectiveness of the proposed FAGS module, we conducted ablation experiments. As depicted in Figure \ref{fig:ame_ablation_quali}, the FAGS with $L_g$ significantly enhances the visual quality of the generated samples. The method of RSSA \cite{xiao_few_2022} with self-correlation consistency loss $L_{scc}$, which removes the pre-trained generative model, use the training images directly as the source domain and transfer their information to the target generator. However, due to the limited information in the few-shot source domain, the method leads to blurry images. We address these problems by introducing the FAGS module to create a pseudo-source domain by building the Geodesic surface. However, if we replace the loss function from $L_{scc}$ to others, like the smooth-$\ell$1 loss, it results in inadequate fidelity. Similar conclusions can be drawn from the quantitative results presented in Table \ref{table:ablation}. 

\subsubsection{Effect of the I$\&$R module}
\label{sec:Effect of the InSR}

\begin{figure}[htbp]
\begin{center}
\includegraphics[width=0.8\linewidth]{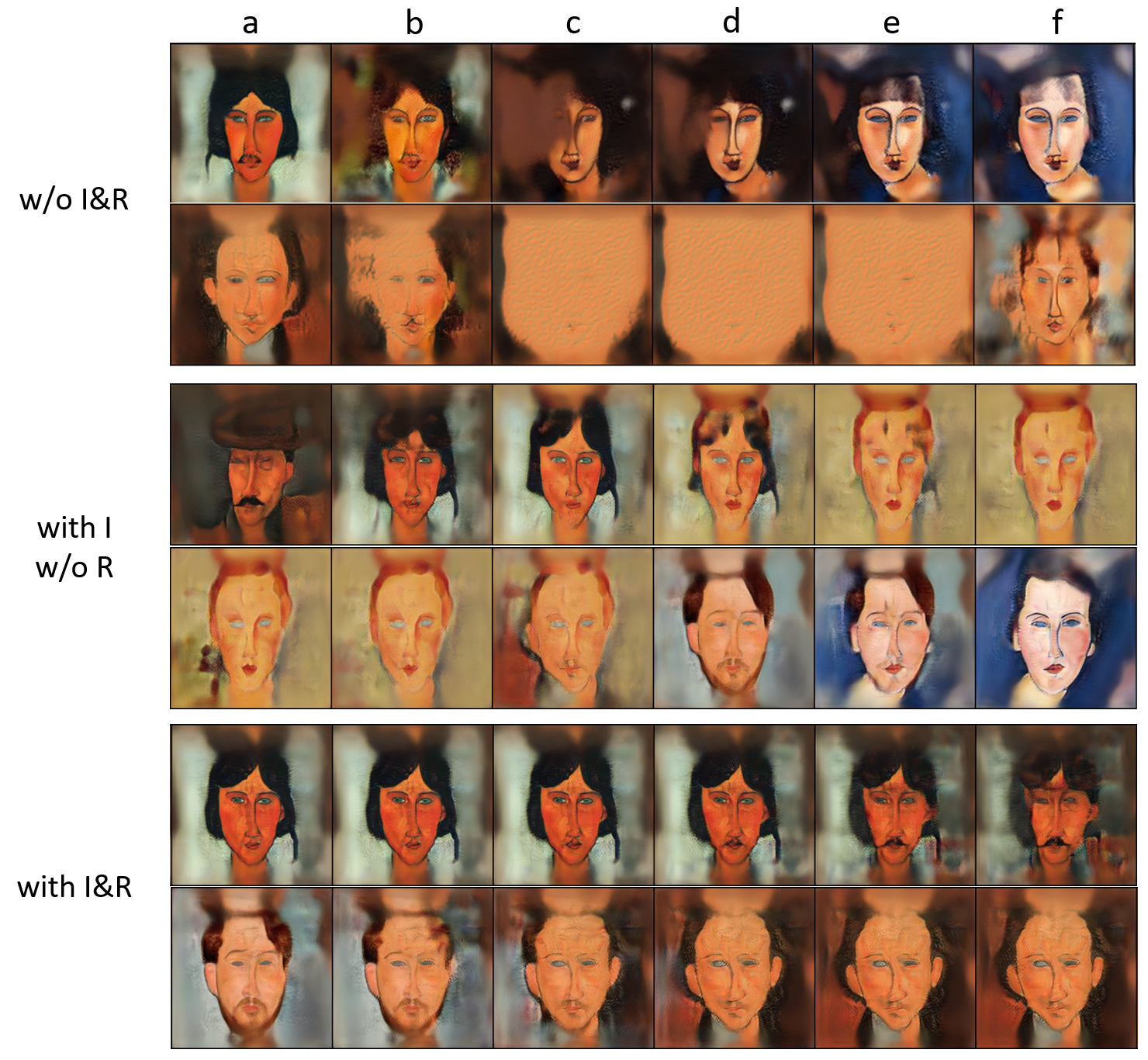}
\end{center}
\caption{Qualitative ablation on I\&R module.}
\label{fig:ame_ablation_InSR_quali}
\end{figure}

Figure \ref{fig:ame_ablation_InSR_quali} illustrates the impact of including or excluding the I\&R module on the generation of interpolated images in our model.
If I\&R is not applied, significant blurriness is presented in the intermediate interpolated samples, as shown in Figure \ref{fig:interp} and the first two rows of Figure \ref{fig:ame_ablation_InSR_quali} from column b to column e. If the $L_{inp}$ is introduced for the interpolated images, the fogging phenomenon is completely eliminated, as shown in the third and fourth rows of Figure \ref{fig:ame_ablation_InSR_quali}. If the $L_{dr}$ is removed, the ``stairlike" phenomenon occurs by observing the transition from column a to column b and from column c to column d in the third and fourth rows of Figure \ref{fig:ame_ablation_InSR_quali}, respectively. The fifth and sixth rows show the generation with full I\&R module, which has the best visual quality. Thus, the Regularization strategy in the I\&R module can alleviate the “stairlike” phenomenon.


\section{Conclusion}
\label{sec:Conclusion}

In this paper, we propose Information Transfer from the Built Geodesic Surface (ITBGS), which transfer information without the readily source domain. ITBGS contains two modules: Feature Augmentation on Geodesic Surface (FAGS);  Interpolation and Regularization (I\&R). With the FAGS module, a Geodesic surface, i.e., a pseudo-source domain, is built in the Pre-Shape space. From the pseudo-source domain, the adaption methods can preserve and transfer the inherent information to the target domain. I\&R module supervise and regularize the interpolated images and their relative distances. Thus, the synthesis performance of the generative model can be enhanced with the I\&R module.

The proposed ITBGS also has some limitations. The generated images of ITBGS primarily fuse visual elements from the training set seamlessly. However, models trained with ITBGS is not capable of modifying these visual elements or generating entirely new ones. We anticipate the development of more powerful and data-efficient models that will generate higher quality images and significantly contribute to various downstream tasks such as few-shot image segmentation and recognition.

\section*{Acknowledgments}
This research is sponsored by National Natural Science Foundation of China (Grant No. 52273228), Key Research Project of Zhejiang Laboratory (No. 2021PE0AC02), Key Program of Science and Technology of Yunnan Province (202302AB080022), the Project of Key Laboratory of Silicate Cultural Relics Conservation (Shanghai University), Ministry of Education (No. SCRC2023ZZ07TS).

\bibliographystyle{IEEEtran}
\bibliography{few_shot_image_generation.bib}

\end{document}